\documentclass[conference]{IEEEtran}
\IEEEoverridecommandlockouts
\usepackage{cite}
\usepackage{amsmath,amssymb,amsfonts}
\usepackage{algorithmic}
\usepackage{graphicx}
\usepackage{textcomp}
\usepackage{xcolor}
\def\BibTeX{{\rm B\kern-.05em{\sc i\kern-.025em b}\kern-.08em
    T\kern-.1667em\lower.7ex\hbox{E}\kern-.125emX}}
\begin{document}

\title{Bridging Industrial Expertise and XR with LLM-Powered Conversational Agents}

\author{
\IEEEauthorblockN{Despina Tomkou\IEEEauthorrefmark{1},
George Fatouros\IEEEauthorrefmark{1}\IEEEauthorrefmark{3},
Andreas Andreou\IEEEauthorrefmark{2},
Georgios Makridis\IEEEauthorrefmark{3}\\
Fotis Liarokapis\IEEEauthorrefmark{2},
Dimitrios Dardanis \IEEEauthorrefmark{3},
Athanasios Kiourtis \IEEEauthorrefmark{3},
John Soldatos\IEEEauthorrefmark{1},
Dimosthenis Kyriazis\IEEEauthorrefmark{3}}
\IEEEauthorblockA{\IEEEauthorrefmark{1}\textit{Innov-Acts Ltd.},
Nicosia, Cyprus\\
\{dtomkou, gfatouros, jsoldat\}@innov-acts.com}
\IEEEauthorblockA{\IEEEauthorrefmark{2}
\textit{CYENS Centre of Excellence},
Nicosia, Cyprus\\
\{a.andreou,  f.liarokapis\}@cyens.org.cy}
\IEEEauthorblockA{\IEEEauthorrefmark{3}
\textit{University of Piraeus},
Piraeus, Greece\\
\{gfatouros, gmakridis, ddardanis, kiourtis, dimos\}@unipi.gr}
}

\maketitle

\begin{abstract}
This paper introduces a novel integration of Retrieval-Augmented Generation (RAG) enhanced Large Language Models (LLMs) with Extended Reality (XR) technologies to address knowledge transfer challenges in industrial environments. The proposed system embeds domain-specific industrial knowledge into XR environments through a natural language interface, enabling hands-free, context-aware expert guidance for workers. We present the architecture of the proposed system consisting of an LLM Chat Engine with dynamic tool orchestration and an XR application featuring voice-driven interaction. Performance evaluation of various chunking strategies, embedding models, and vector databases reveals that semantic chunking, balanced embedding models, and efficient vector stores deliver optimal performance for industrial knowledge retrieval. The system's potential is demonstrated through early implementation in multiple industrial use cases, including robotic assembly, smart infrastructure maintenance, and aerospace component servicing. Results indicate potential for enhancing training efficiency, remote assistance capabilities, and operational guidance in alignment with Industry 5.0's human-centric and resilient approach to industrial development.
\end{abstract}


\section{Introduction}
\label{sec:1}

Industry 5.0 is redefining industrial operations by emphasizing sustainable, resilient and human-centric systems \cite{rovzanec2023human}. It recognizes the role of industry not only in economic growth but also in achieving broader societal objectives by prioritizing worker well-being. At the same time, it advocates for the development of agile business processes capable of withstanding geopolitical disruptions and natural crises \cite{europeancommission_industry_2021}.

Despite rapid technological progress, industries still face critical challenges related to knowledge transfer and the availability of expert support \cite{wurster_large_2024}. A shortage of skilled professionals, combined with geographical constraints that often necessitate on-site presence, results in significant production bottlenecks across various sectors \cite{ongbali2021study}. For example, when Samsung relocated one of its major production facilities, the company encountered a substantial decline in skilled labor due to difficulties in retaining and attracting specialized personnel. As a result, experts had to be flown in from other regions for essential maintenance and operational support, leading to costly downtime and logistical complications \cite{article_samsung}. This case illustrates the pressing need for efficient, remotely accessible knowledge-sharing solutions—particularly in scenarios involving maintenance, assembly and troubleshooting, where timely professional intervention is crucial.

Extended Reality (XR) technologies are increasingly used in industrial contexts to support training, problem-solving and remote assistance. By offering immersive simulations of real-world environments and equipment, XR enables safe, hands-on instruction without requiring access to physical machinery and allows experts to guide workers remotely. However, many existing XR applications operate in isolation from domain-specific data, which limits their effectiveness—especially in the absence of on-demand expert availability.

These limitations are evident across several core use cases:
\begin{itemize}
    \item Training: XR provides safe, naturalistic environments for practicing complex tasks without risking damage to real equipment or disrupting operations. However, engagement is typically restricted to predefined teaching modules that cannot easily adapt to the specific needs or experience of individual trainees.
    \item Remote Assistance: XR enables off-site experts to provide visual guidance during on-site operations but lacks deep integration with structured industrial documentation.
    \item Real-Time Assistance: Augmented Reality (AR) overlays can guide workers step-by-step in real-time, yet often miss the dynamic reasoning and flexibility required in difficult, unexpected situations.
\end{itemize}

The convergence of Large Language Models (LLMs) and XR technologies presents a promising avenue to overcome these limitations. Recent advancements in generative AI—exemplified by models such as OpenAI’s GPT-4—have demonstrated strong capabilities in natural language understanding, generation and question answering. With appropriate prompt engineering, these models can be tailored to user-specific needs \cite{fatouros_transforming_2023}. Furthermore, by incorporating Retrieval-Augmented Generation (RAG) techniques \cite{lewis2020retrieval}, LLMs can access and utilize external data sources, thereby improving factual accuracy and enabling domain-specific expertise \cite{fatouros2025marketsenseai}. When combined, the immersive capabilities of XR and the contextual intelligence of RAG-enhanced LLMs can yield advanced support systems for training and operational guidance that are both interactive and adaptable to individual users' proficiency and learning pace.

This paper introduces a novel system that embeds domain-specific industrial knowledge into XR environments via a RAG-enhanced LLM-powered conversational engine. At its core, the system incorporates technical documentation into XR settings through a natural language interface. To ensure high-quality retrieval performance, we evaluate various chunking strategies, embedding models and vector databases to optimize the relevance and speed of information access. An implementation framework is also developed, enabling voice-based interaction with this embedded knowledge in XR environments.

The system is applied in multiple industrial scenarios, including robotic assembly, smart water pipe maintenance, aircraft component servicing and edge device installation \cite{kiourtis2024xr5}. These applications demonstrate how the proposed approach enhances XR-based support with real-time, contextual and personalized knowledge delivery, supporting a more efficient and human-centric collaboration.

The remainder of this paper is organized as follows: Section~\ref{sec:2} reviews related work on LLM-powered conversational assistants and XR applications in industry. Section~\ref{sec:3} presents the system architecture along with preliminary evaluation results. Section~\ref{sec:4} discusses real-world use cases and implementations. Section~\ref{sec:5} concludes the paper with key insights and directions for future research.

\section{Background and Related Work}
\label{sec:2}

\subsection{LLMs as Conversational Assistants}
The integration of LLMs with XR technologies can reshape industrial and manufacturing applications by providing intelligent, context-aware conversational interfaces. Transformer-based models such as GPT-4, Claude, Gemini, and LLaMA have proven highly effective across diverse AI applications, including code generation, search enhancement and decision-making support due to their large-scale pretraining and zero-shot generalization capabilities \cite{shao_survey_2024, fatouros2024can}.

However, despite these strengths, LLMs exhibit critical limitations when applied to domain-specific contexts like industrial training and maintenance. Issues such as hallucinations, outdated or imprecise responses, and poor adaptability can pose significant risks in precision-critical industrial environments \cite{myohanen2023improving}. To address these challenges, domain-specific knowledge injection strategies have been proposed, categorized into static embeddings, modular adapters, prompt engineering, and dynamic retrieval \cite{song_injecting_2025}. Among these, dynamic retrieval—implemented through RAG is recognized as particularly effective in evolving knowledge domains, allowing LLMs to dynamically access external documentation and manuals through vector databases.

Recent advancements, such as Agentic RAG, further enhance dynamic retrieval by employing autonomous AI agents capable of advanced reasoning, contextual memory management, and adaptive retrieval mechanisms \cite{singh2025agenticrag}. These agents break down complex user queries into actionable steps, thus providing consistent, personalized and accurate guidance \cite{bousetouane2025agenticsystems}. Real-world deployments, including an LLM-based assistant for training supply chain employees, have demonstrated significant cost reductions and improved user satisfaction through the provision of immediate and clear instructions.
\cite{gezdur_innovators_2025}.

In manufacturing, LLMs can address labor shortages and skill gaps by providing interactive mentoring, real-time assistance, predictive maintenance, and regulatory support \cite{inbook}. These capabilities enable less experienced workers to perform complicated tasks proficiently and safely, highlighting the strategic role of LLMs in industrial settings.

\subsection{AR and VR in Industrial Applications}
AR and Virtual Reality (VR) significantly enhance industrial applications by improving worker interaction with equipment and engagement during operations. Despite their potential, the adoption of these technologies remains limited because of high development costs and the absence of standardized platforms.\cite{wang2020enhancing, moro2021virtual}.

Nevertheless, AR/VR technologies substantially enhance user participation and intuitive interaction with sophisticated industrial systems. Applications such as digital twins, remote collaboration, custom manufacturing, and virtual training scenarios demonstrate considerable value in improving the learning experience and operational throughput. \cite{breque2021industry, carmigniani2011augmented}.

Specifically, AR and VR simplify advanced tasks such as assembly, error detection, and quality assurance through visual guidance and real-time feedback. Designers and engineers utilize these technologies to rapidly prototype products in three-dimensional virtual environments, accelerating product development cycles \cite{berni2020applications}. Training simulations offered by AR/VR platforms enable workers to perform intricate procedures safely in controlled virtual settings. This allows for repeated practice, leading to deeper comprehension and improved task proficiency.\cite{makransky2018structural, marner2013improving}.

Furthermore, XR effectively supports remote assistance by enabling geographically dispersed professionals and field workers to collaborate and interact with equipment in virtual real-time settings. \cite{wang2016comprehensive}.  As a result, travel needs are reduced and response times for expert interventions are improved.

Building upon this existing research, our work integrates RAG-enhanced LLM conversational agents directly into XR environments. Unlike traditional text-based assistants, our system leverages XR interfaces, delivering hands-free, visual, and procedural guidance tailored to user expertise and needs. 

\section{Architecture}
\label{sec:3}
\begin{figure}[htbp]
\centerline{\includegraphics[width=0.5\textwidth]{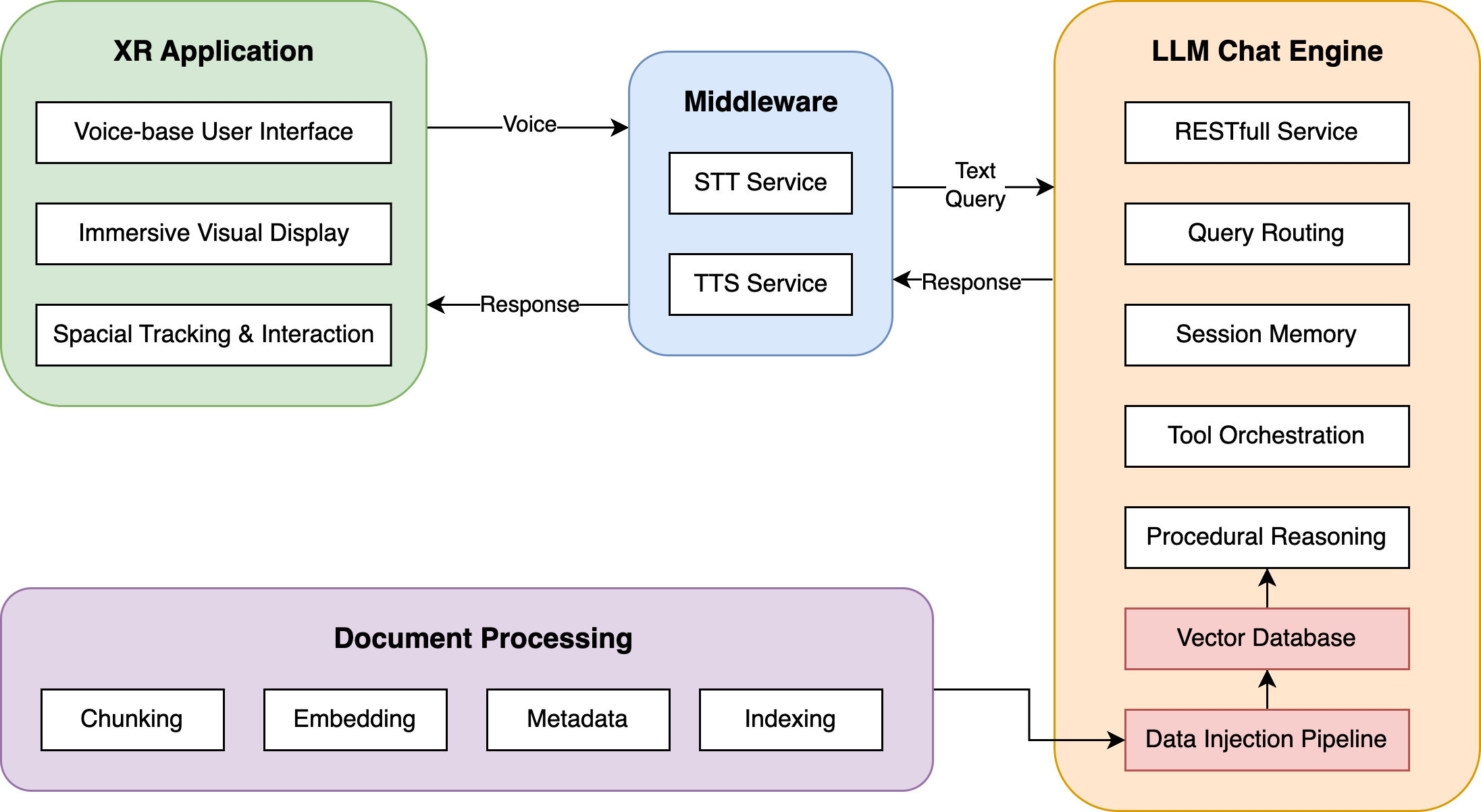}}
\caption{Architectural Overview of RAG-Enhanced LLM with XR Integration for Industrial Environments. The diagram illustrates the bi-directional communication flow between the XR Application and LLM Chat Engine through middleware services, highlighting the document processing pipeline that populates the vector database for knowledge retrieval.}
\label{fig:integration}
\end{figure}

The proposed system integrates RAG-enhanced LLMs with XR to deliver domain-specific knowledge within industrial environments as illustrated in Fig.~\ref{fig:integration}. 

Our architecture delivers expert guidance accessible from voice commands through two primary components: the \emph{LLM Chat Engine} for natural language processing and knowledge retrieval and the \emph{XR Application} for providing immersive visual interfaces. The system operates via a structured communication protocol that connects natural speech recognition in XR to the back-end knowledge services. Through intelligent query routing and RAG techniques, domain-specific information is retrieved from a vector database populated with embedded documentation. The following subsections detail the technical specifications and implementation of each architectural component.

\subsection{LLM Chat Engine}

\begin{figure}[htbp]
\centerline{\includegraphics[width=0.5\textwidth]{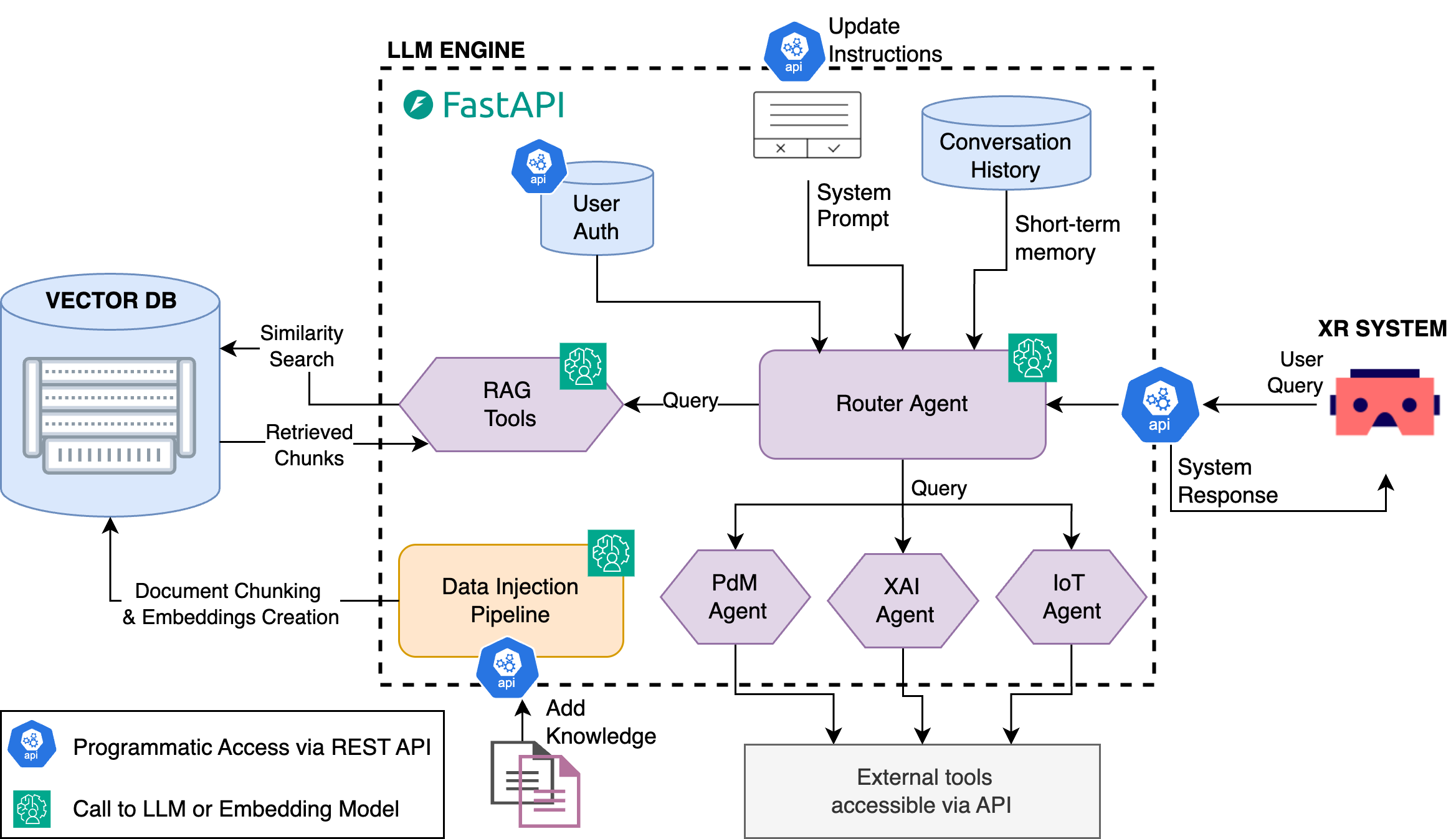}}
\caption{High-level Architectural View of LLM Chat Engine }
\label{fig:arch}
\end{figure}

The LLM Chat Engine forms the core intelligence of the system, orchestrating communication between agents, managing prompts, and maintaining conversation context during industrial XR-based interactions. As illustrated in Fig.~\ref{fig:arch}, the engine integrates a modular architecture centered on the Query Router Agent, which controls the end-to-end query processing workflow. This agent evaluates user inputs, leverages metadata from registered tools, and dynamically routes queries to specialized components based on factors such as query content, historical interactions, and predefined system prompts \cite{ayyamperumal2024current}. These prompts enforce domain-specific constraints, ensuring responses remain aligned with industrial objectives. The following paragraphs elaborate on the LLM Chat Engine's internal components and processes.

\subsubsection{Dynamic Tool Orchestration} Each RAG tool in the engine is associated with a document stored in the vector database. During document ingestion, metadata (e.g., title, version, summary) is registered with the Query Router Agent, enabling targeted retrieval by narrowing the search space. This approach improves efficiency and accuracy compared to brute-force similarity searches across all documents. Additionally, the router invokes auxiliary agents to enhance context-aware responses:

\begin{itemize} 
    \item \textit{PdM Agent:} Fetches predictive maintenance (PdM) outputs from external services, providing insights on machinery condition. 
    \item \textit{XAI Agent:} Retrieves model explanations from an eXplainable AI service, helping the user interpret algorithmic predictions \cite{makridis2025virtualxai}. 
    \item \textit{IoT Agent:} Acquires current and historical sensor readings from IoT devices, enabling real-time monitoring and setup guidance. 
\end{itemize}

\subsubsection{Response Generation} The engine synthesizes answers by combining retrieved document excerpts, system prompts, user history, and outputs from auxiliary agents. This enables automated decomposition of complex industrial queries into actionable steps, functioning as a reasoning engine for real-time decision support.

\subsubsection{Implementation} The engine is deployed as a REST API via FastAPI, offering endpoints for query processing, document injection, and session management. Security is enforced through API key authentication, while LlamaIndex and LangChain frameworks establish agent and tool construction. Session persistence ensures continuity during extended XR training scenarios.

\subsubsection{Knowledge Integration} 
\begin{figure}[htbp] \centerline{\includegraphics[width=0.5\textwidth]{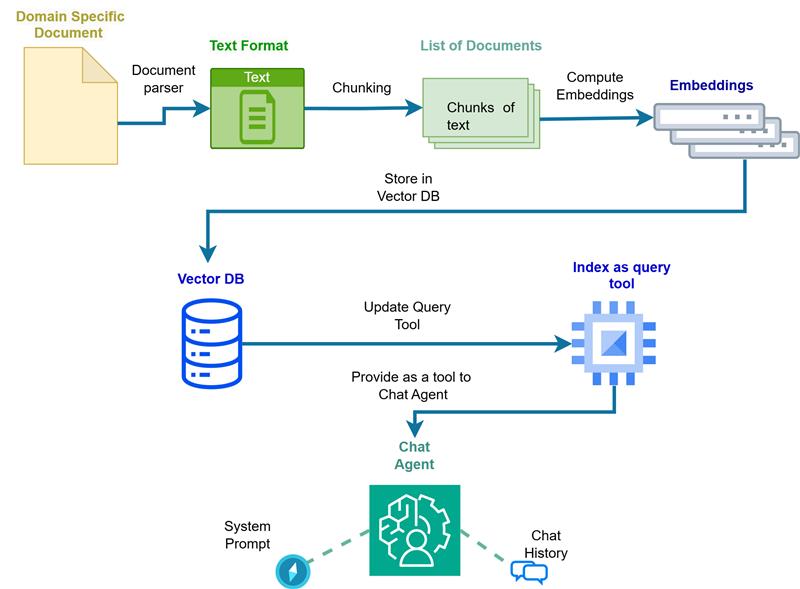}} \caption{Data and Knowledge Injection Pipeline} \label{fig2} \end{figure}

A key sub-system of the LLM Chat Engine is the \emph{Data Injection Component}, which processes and indexes source documents to enable efficient RAG operations. As illustrated in Fig.~\ref{fig2}, the pipeline consists of the following stages: 

\begin{itemize} 
    \item \textit{Document Parsing \& Chunking:}  Three documents—ranging from 74 to 554 pages, with a mean of 331 pages and standard deviation of 197—were  converted to text and then segmented into coherent chunks. Instead of fixed-length splits—which risk incomplete or arbitrary divisions—the system applies a semantic approach that respects headings and subheadings, preserving the logical structure of technical materials. 
    \item \textit{Embeddings Creation:} Each chunk is transformed into a high-dimensional vector representation, capturing semantic content. This facilitates robust similarity matching even when queries do not use the exact wording of the original text. 
    \item \textit{Vector Database:} The resulting vectors are stored in a database that supports efficient similarity search. Metadata are crucial here, allowing the system to filter out irrelevant content before performing vector retrieval.
    \item \textit{Query Tools:} One specialized query tool per use case leverages both embeddings and metadata for targeted retrieval. Each tool includes specific fields such as author, document type, and version to refine the scope of a search. 
\end{itemize}

\subsubsection{Performance Evaluation} We benchmarked the retrieval system using RAGChecker \cite{ru2024ragchecker}, testing three recognized strategies for chunking, embedding models, and vector databases on industrial documents of varying lengths and domains. Each approach was evaluated against four metrics: \emph{Claim Recall (CR)}, \emph{Context Precision (CP)}, \emph{Hallucination (Hallu.)}, and \emph{Faithfulness (Faith.)}. Tables~\ref{tab:chunking_strategies}--\ref{tab:vector_stores} summarize the top-performing configurations.

Among the chunking strategies, Semantic Chunking outperformed fixed-length methods, reaching a faithfulness of 99.07\% while minimizing hallucination (Table~\ref{tab:chunking_strategies}). For embeddings, the Mpnet model yielded the highest recall (95.28\%) but had greater hallucination, whereas OpenAI’s model balanced both accuracy and reliability (Table~\ref{tab:embedding_strategies}) \cite{song_mpnet_2020, korade2024strengthening}. Finally, Pinecone ranked highest in the vector database comparison, achieving 93.37\% recall and 96.5\% faithfulness (Table~\ref{tab:vector_stores}). These results validate the engine’s ability to provide precise, context-aware technical guidance for industrial XR applications.

\begin{table}[htbp]
\caption{Chunking Strategy Performance Metrics}
\label{tab:chunking_strategies}
\centering
\begin{tabular}{|l|c|c|c|c|c|}
\hline
\textbf{Chunking} & \textbf{CR} & \textbf{CP} & \textbf{Hallu.} & \textbf{Faith.} & \textbf{Rank} \\
\hline
Semantic Context & 93.05 & \textbf{99.13} & \textbf{0.21} & \textbf{99.07} & 1 \\
\hline
Fixed length=2028 & \textbf{97.12} & 99.05 & 3.59 & 91.14 & 2 \\
\hline
Fixed length=1024 & 85.41 & 95.37 & 4.51 & 95.49 & 3 \\
\hline
\end{tabular}
\end{table}

\begin{table}[htbp]
\caption{Embedding Model Performance Metrics}
\label{tab:embedding_strategies}
\centering
\begin{tabular}{|l|c|c|c|c|c|}
\hline
 \textbf{Embedding} & \textbf{CR} & \textbf{CP} & \textbf{Hallu.} & \textbf{Faith.} & \textbf{Rank} \\
\hline
Mpnet & \textbf{95.28} & \textbf{99.4} & 5.37 & 90.28 & 1 \\
\hline
OpenAI - small & 86.61 & 96.47 & \textbf{1.06} & 98.97 & 2 \\
\hline
OpenAI - ada & 93.7 & 98.6 & 3.56 & \textbf{96.49} & 3 \\
\hline
\end{tabular}
\end{table}

\begin{table}[htbp]
\caption{Vector Store Performance Metrics}
\label{tab:vector_stores}
\centering
\begin{tabular}{|l|c|c|c|c|c|}
\hline
\textbf{Vector DB} & \textbf{CR} & \textbf{CP} & \textbf{Hallu.} & \textbf{Faith.} & \textbf{Rank} \\
\hline
Pinecone & \textbf{93.37} & 97.22 & 2.54 & \textbf{96.5} & \textbf{1} \\
\hline
Chroma & 92.11 & \textbf{98.05} & \textbf{2.01} & 95.18 & 2 \\
\hline
Faiss & 90.11 & \textbf{98.05} & 3.21 & 94.48 & 3 \\
\hline
\multicolumn{6}{l}{\textit{Note}: Bold values indicate the best performance for each metric.}
\end{tabular}
\end{table}

\subsection{XR Application}

Recent advancements in XR have revolutionized human-computer interaction, particularly through the integration of sophisticated speech processing capabilities. In our research, we propose a novel speech interaction framework for augmented reality applications that leverages AI to create intuitive, hands-free user experiences.

\begin{figure}[htbp]
\centerline{\includegraphics[width=0.5\textwidth]{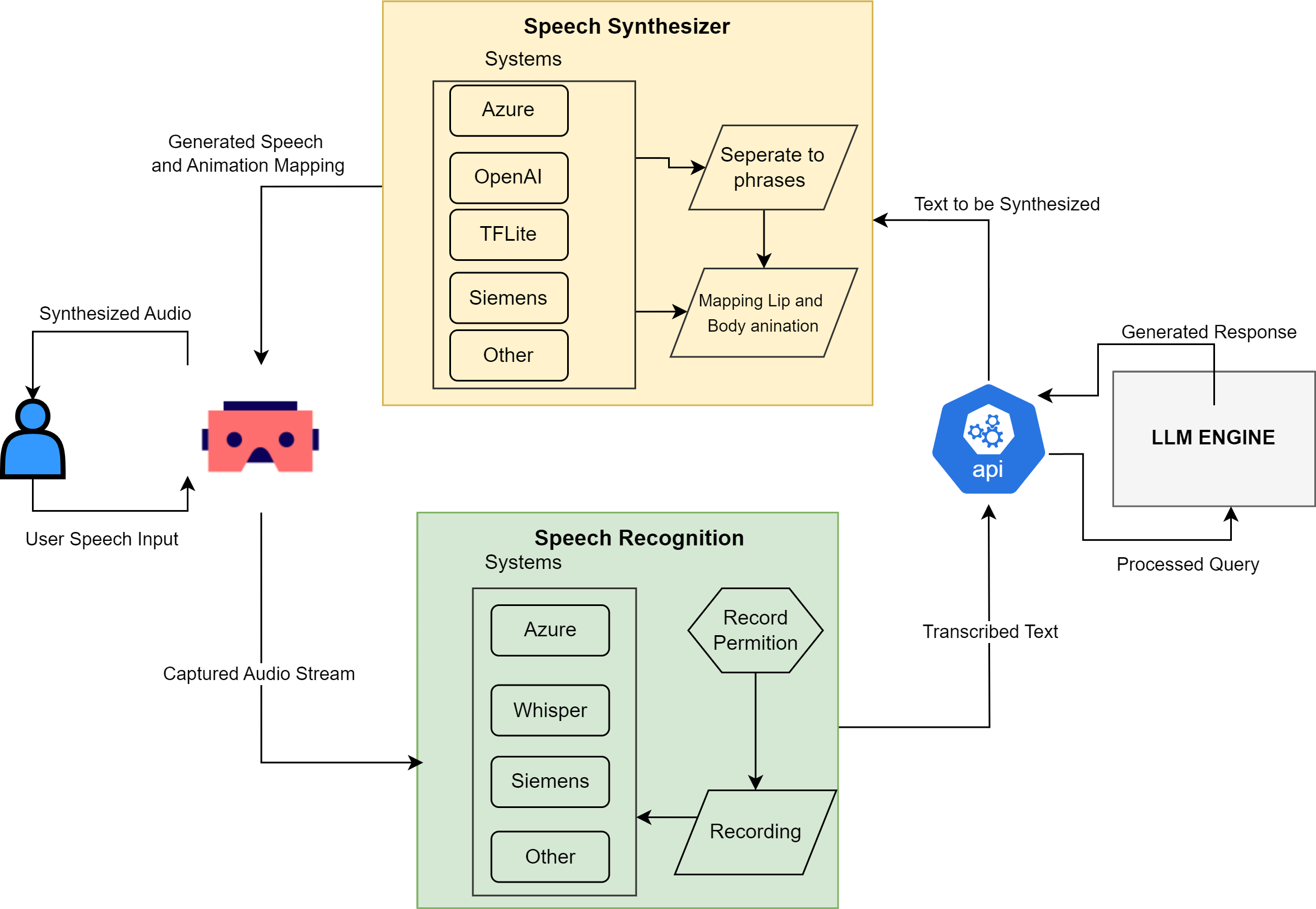}}
\caption{ XR Architecture }
\label{fig:xr}
\end{figure}

Fig.~\ref{fig:xr} illustrates the speech interaction architecture in an AR application, integrating Speech-to-Text (STT) and Text-to-Speech (TTS) systems to enable seamless communication with the LLM Chat Engine. The process begins when the user speaks into the AR app, which captures the audio and forwards it to the speech recognition system. The STT module, powered by engines such as Azure Speech, Whisper AI, or Siemens AI Voice Assistant, transcribes the speech into text and sends it to the Engine's API.

Once the LLM Chat Engine processes the user’s query, it generates a text response. This response is sent back to the speech synthesizer, which converts the text into natural-sounding speech using TTS technologies like Azure Speech, OpenAI TTS, or TFLite models. Additionally, the system includes a mechanism for mapping speech to lip and body animations, ensuring a more interactive AR experience.

Real-time processing capabilities of STT and TTS technologies reduce latency, enabling fluid and natural conversations. This bidirectional speech interaction eliminates the need for manual input, making hands-free AI chatbot interactions possible. Furthermore, the integration of deep-learning models enhances noise reduction and supports multiple languages, improving accessibility and usability across different environments.

This architecture allows users to communicate effortlessly with the LLM Chat Engine, making applications more interactive and engaging. Whether for remote AI assistance, interactive learning, or accessibility support, the combination of STT and TTS in XR enhances user experience by providing audio feedback and visual interactions in real-time.

\section{Application and Use Cases}
\label{sec:4}

The proposed LLM-XR architecture is being evaluated across different industrial environments, each presenting unique challenges in training, remote assistance and operational efficiency. These environments are used to explore the feasibility of the proposed solution through early-stage evaluation and testing

\subsection{Support for Industrial Assembly and Maintenance} 
\begin{figure}[htbp]
\centerline{\includegraphics[width=0.5\textwidth]{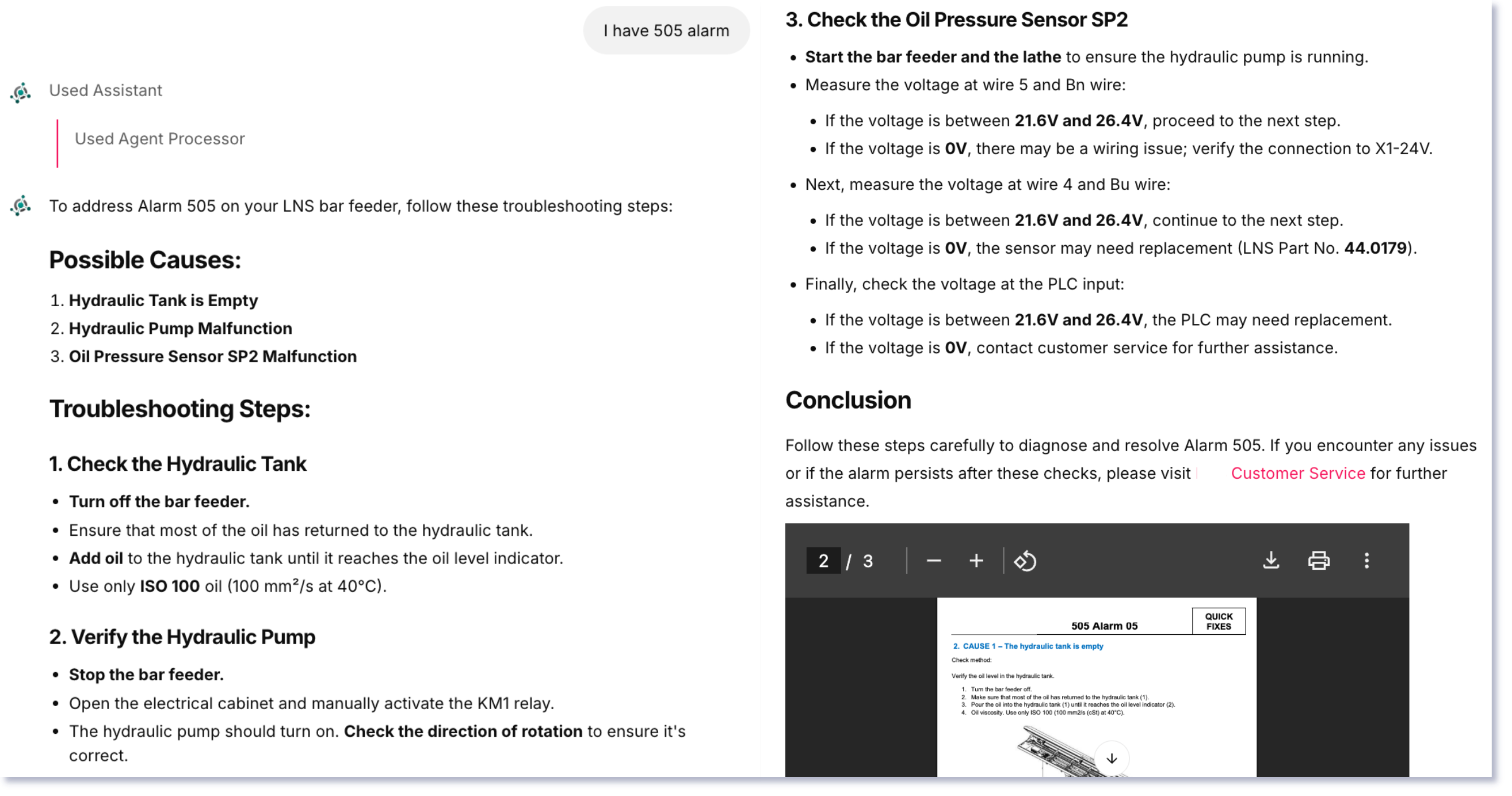}}
\caption{Example response from the LLM Chat Engine to a user query.}
\label{fig:lns}
\end{figure}

In manufacturing environments that involve robotic assembly and system commissioning, our system is being integrated into XR workflows in experimental setups to assess its potential to support and improve remote troubleshooting. In virtual training modules, technicians get an engaging way to learn, thus reducing the need for physical testing. The proposed system supports hands-free access to commissioning checklists, troubleshooting guides and equipment manuals, enabling novice and expert users to reduce downtime, errors and enhance productivity. Fig.~\ref{fig:lns} depicts an example response from LLM Chat Engine to a user query requesting assistance in troubleshooting citing the source document.

\subsection{Predictive Maintenance in Smart Infrastructure Operations}
\begin{figure}[htbp]
\centerline{\includegraphics[width=0.5\textwidth]{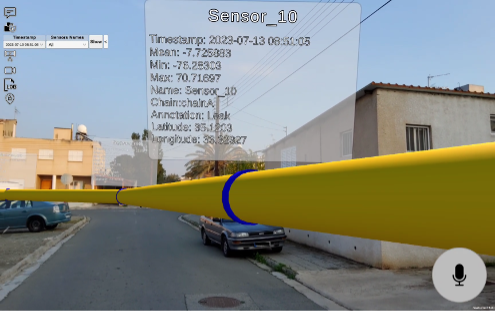}}
\caption{AR application of a smart water pipe rendering IoT data available from LLM agent.}
\label{fig:ar-iot}
\end{figure}

\begin{figure}[htbp]
\centerline{\includegraphics[width=0.5\textwidth]{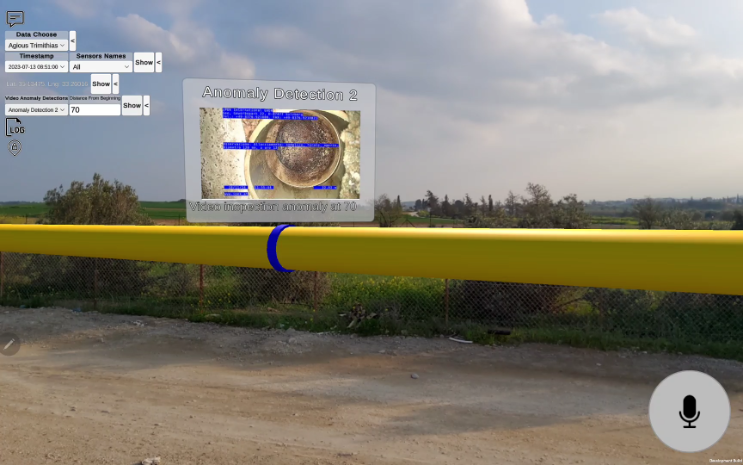}}
\caption{AR application of a smart water pipe rendering predictive maintenance data available from LLM agent.}
\label{fig:ar-pdm}
\end{figure}

Our solution supports predictive maintenance for infrastructure scenarios involving complex systems, such as smart water networks. The system is being designed to provide simulations driven by anomaly detection models, with content dynamically adjusted from LLM Chat engine's  agents utilizing external services and tools. Visual overlays can show malfunction indicators, sensor data and AI-generated recommendations such as step-by-step instructions offering granular support during inspections and repair workflows. Fig.~\ref{fig:ar-iot} and \ref{fig:ar-pdm} illustrate an example AR application of a smart water pipe assisting technicians to find the location of a leakage.

\subsection{Stress-aware Maintenance in Safety-Critical Environments}
Our XR-LLM integration focuses on safety, precision and stress mitigation in aerospace contexts where component maintenance involves high cognitive and physical demand. The assistant can provide step-by-step safety procedures and answer real-time maintenance questions, aims to provide feedback and reference while considering biometric from wearables such as smartwatches.

\section{Conclusion and Future Work}
\label{sec:5}

Our paper has presented a novel architecture integrating LLM agents with XR environments to address critical knowledge transfer challenges in industrial settings. By combining an intelligent LLM Chat Engine with XR applications, our system enables hands-free, context-aware interactions that adapt to users' expertise levels across diverse industrial scenarios. Performance evaluation of various chunking strategies, embedding models, and vector databases has validated our approach to knowledge retrieval, with semantic chunking, balanced embedding models, and efficient vector stores like Pinecone delivering optimal results for industrial documentation.
Future work will focus on full-scale integration and evaluation of our system in the presented industrial use cases, implementing required optimizations based on deployment feedback. We plan to test the LLM Chat Engine with smaller, locally-hosted open-source models to address privacy concerns when handling confidential documents. Additionally, we aim to enhance multimodal capabilities through computer vision integration and implement advanced personalization based on user performance and biometric feedback. This continued development promises to further bridge the gap between immersive technologies and contextual intelligence for industrial applications.

\section*{Acknowledgment}

Part of the research leading to the results presented in this paper has received funding from the European Union’s funded Project XR5.0 [GA 101135209].

\bibliography{references}
\bibliographystyle{IEEEtran}

\end{document}